\documentclass[10pt,twocolumn,letterpaper]{article}

\usepackage[pagenumbers]{iccv} 

\definecolor{iccvblue}{rgb}{0.21,0.49,0.74}
\usepackage[pagebackref,breaklinks,colorlinks,allcolors=iccvblue]{hyperref}

\def\paperID{*****} 
\def\confName{ICCV}
\def\confYear{2025}

\title{Bootstrapping Physics-Grounded Video Generation through \\ VLM-Guided Iterative Self-Refinement}

\author{
	Yang Liu\textsuperscript{1}\hspace{2em} Xilin Zhao\textsuperscript{2} \hspace{2em} Peisong Wen\textsuperscript{1}\hspace{2em} Siran Dai\textsuperscript{3,4}\hspace{2em} Qingming Huang\textsuperscript{1,5}\thanks{Corresponding authors} \\
	{\textsuperscript{1}School of Computer Science and Technology, University of Chinese Academy of Sciences} \\
    {\textsuperscript{2}School of Computer Science and Technology, Beijing Institute of Technology} \\
    {\textsuperscript{3}Institute of Information Engineering, Chinese Academy of Sciences} \\
    {\textsuperscript{4}School of Cyber Security, University of Chinese Academy of Sciences} \\
    {\textsuperscript{5}Institute of Computing Technology, Chinese Academy of Sciences} \\
	{\tt\small liuyang232@mails.ucas.ac.cn\hspace{2em} 13426118680@163.com\hspace{2em} wenpeisong@ucas.ac.cn} \\ 
    {\tt\small  daisiran@iie.ac.cn\hspace{2em} qmhuang@ucas.ac.cn }
}

\begin{document}
\maketitle

\begin{abstract}
Recent progress in video generation has led to impressive visual quality, yet current models still struggle to produce results that align with real-world physical principles.
To this end, we propose an iterative self-refinement framework that leverages large language models and vision-language models to provide physics-aware guidance for video generation.
Specifically, we introduce a multimodal chain-of-thought (\textbf{MM-CoT}) process that refines prompts based on feedback from physical inconsistencies, progressively enhancing generation quality.
This method is training-free and plug-and-play, making it readily applicable to a wide range of video generation models.
Experiments on the PhyIQ benchmark show that our method improves the Physics-IQ score from 56.31 to \textbf{62.38}.
We hope this work serves as a preliminary exploration of physics-consistent video generation and may offer insights for future research.
\end{abstract}    

\section{Introduction}
\label{sec:introduction}

Recent advances in video generation have led to remarkable progress, exemplified by models such as Sora~\cite{Sora}, Lumiere~\cite{Lumiere}, and VideoPoet~\cite{VideoPoet}, which produce high-quality videos with clear details, natural dynamics, and realistic rendering.
However, the gap between current video generation systems and world models remains substantial. A key step toward bridging this gap lies in establishing a stronger connection between generative models and the physical principles of the real world~\cite{challenge1,challenge2,challenge3,challenge4,challenge5}.

With the expansion of training datasets and the scaling up of model capacity, an increasing number of video generation models (VGMs) have demonstrated preliminary abilities in modeling real-world physical dynamics through large-scale pretraining~\cite{pretrain1,pretrain2} or distillation fine-tuning~\cite{distill1,distill2} based on advanced visual representations~\cite{dinov2,dinov3,tcore,wen2025semantic,HAP_VR,dai2025exploring}. However, how to effectively elicit such physics-aware generation ability remains an important challenge.
Prior work has shown that well-designed prompts not only steer models toward desired outputs but can also trigger capability emergence~\cite{prompt1,prompt2,prompt3}. Therefore, we believe that high-quality prompt design is essential for guiding video models to generate physics-grounded content.

Concurrently, large language models (LLMs)~\cite{GPT3,GPT4,Deepseek} and vision-language models (VLMs)~\cite{Gemini,Qwen} have achieved rapid progress, delivering breakthroughs across vision, language, and cross-modal tasks.
These advances provide a strong foundation for the automated construction of physics-aware prompts.
Building on this, we leverage state-of-the-art LLMs and VLMs in an iterative physics-guided prompting framework, introducing multimodal chain-of-thought (\textbf{MM-CoT}) reasoning to progressively elicit the video generation model’s physical modeling capabilities.

\begin{figure}[t]
  \centering
    \includegraphics[width=0.98\linewidth]{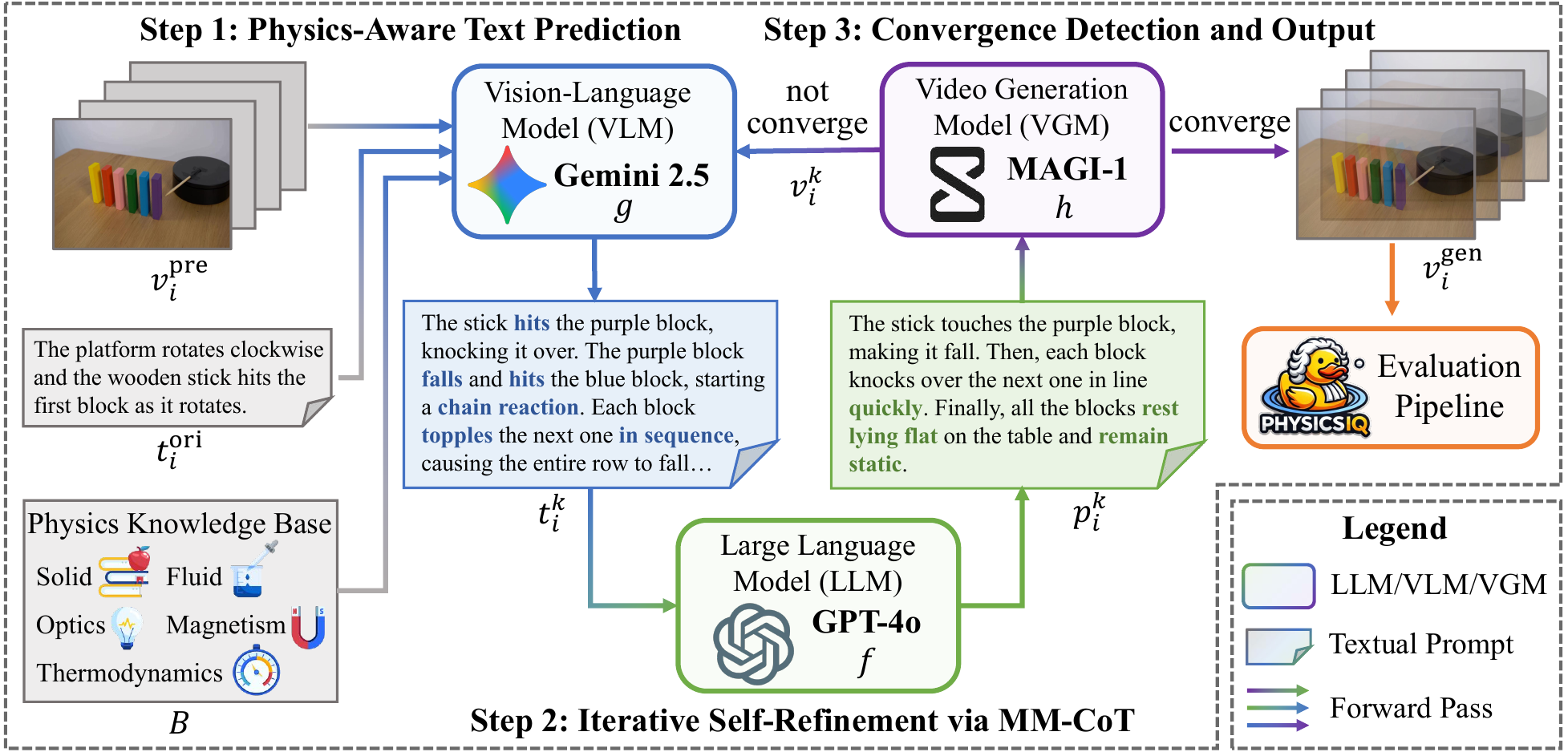}
    \vspace{-8pt}
    \caption{Overview of our method.}
    \vspace{-8pt}
  \label{fig:pipeline}
  \vspace{-12pt}
\end{figure}

Specifically, we first provide the VLM with a concise physics prior, along with the prefix video and its original description from the Challenge. The VLM then generates a detailed prediction of future dynamics, incorporating explicit physical cues.
Subsequently, this prediction is rewritten by an LLM into a prompt compatible with the VGM, which then generates the future segment conditioned on the prefix video.
The generated video is fed back into the VLM to identify potential violations of physical laws, producing a revised description that is then rewritten and fed back into the VGM.
The iterative process continues until the LLM’s output converges, indicating the model’s physics-grounded generation has been sufficiently elicited.

Experiments on the PhyIQ benchmark~\cite{PhyIQ} show that our method achieves a Physics-IQ Score of 62.38, yielding an improvement of 6.07 over the baseline on the leaderboard with a score of 56.31, which suggests its effectiveness in enhancing physical consistency.
Notably, the proposed framework is training-free and plug-and-play, making it readily applicable to a variety of state-of-the-art VGMs.

\section{Method}
\label{sec:method}

This task focuses on generating physically consistent video continuations.
Given a 3-second prefix video $V^{\text{pre}}$ and a short textual description of the scene $T^{\text{ori}}$, the goal is to generate the next 5 seconds of video $V^{\text{gen}}$ in a way that is temporally coherent and physically plausible.

As shown in \cref{fig:pipeline}, we propose a physics-aware video generation framework guided by the collaborative reasoning of LLM and VLM.
The pipeline consists of three stages:

\textbf{Step 1: Physics-Aware Text Prediction.}
Leveraging the VLM's strong capabilities in video understanding and physical reasoning, we first extract explicit physical cues as textual predictions.  
The system input consists of a concise physics knowledge base $B$ and task-specific instructions $I$.  
For each sample, the VLM $f$ processes the 3-second prefix video $v^{\text{pre}}_i$ and its corresponding textual description $t^{\text{ori}}_i$ to generate a detailed, physics-enriched prediction:
$t^{\text{1}}_i = f(t^{\text{ori}}_i, v^{\text{pre}}_i; B, I)$.
However, this output is often verbose and semantically misaligned with the VGM, making it unsuitable as a direct prompt.  
To bridge this gap, we introduce an LLM $g$ to rewrite and simplify the VLM output, producing a concise prompt:
$p^{1}_i = g(t^{\text{1}}_i)$.

\textbf{Step 2: Iterative Self-Refinement via Multimodal Reasoning Chain.}
The simplified prompt $p^{1}_i$ and the prefix video $v^{\text{pre}}_i$ are fed into the VGM $h$ to generate a continuation video $v^{1}_i = h(p^{1}_i, v^{\text{pre}}_i)$.  
As a single pass may not yield physically consistent results, we introduce an iterative refinement loop.  
The generated video is re-processed by the VLM, which detects physical inconsistencies and produces an updated description emphasizing missing or violated physical cues.  
This description is then refined by the LLM into a new prompt, forming a multimodal chain-of-thought across iterations: $p^{k+1}_i = g(f(h(p^{k}_i, v^{\text{pre}}_i); B, I))$.


\textbf{Step 3: Convergence Detection and Output.}
The process continues until the prompts converge, i.e., when $p^{k+1}_i \approx p^{k}_i$, indicating that the VGM’s capacity for modeling the scene’s physical dynamics has been sufficiently activated. 
Finally, the generated video $v^{\text{gen}}_i=v^{k}_i$ is returned.

\section{Experiments}
\label{sec:experiments}

\noindent\textbf{Implementation Details.}
We employ GPT‑4o~\cite{GPT4} as the LLM, Gemini 2.5 Pro~\cite{Gemini} as the VLM, and MAGI‑1~\cite{MAGI1} as the VGM.
The entire pipeline is implemented on the Dify automated workflow platform and a local PyTorch 2.2~\cite{Pytorch} environment.
During inference, we experiment with 16 and 32 steps.
All generated videos are 5 seconds at 24 FPS.

\noindent\textbf{Quantitative Results.}
The evaluation results of our method on the benchmark are presented in \cref{tab:results}.
Overall, iterative prompting leads to consistent performance gains, driven by the VLM’s strong video understanding and the LLM’s ability to generate physics-aware prompts, which progressively activate the VGM’s latent capacity for physical modeling.
Due to varying video complexity, the benefits of iteration do not emerge uniformly across samples.
To mitigate this, we adopt a simple ensemble strategy that combines the best outputs from multiple iterations, achieving a Physics-IQ score of 62.38, which improves upon the baseline by 6.07.

\begin{figure}[t]
  \centering
    \includegraphics[width=0.9\linewidth]{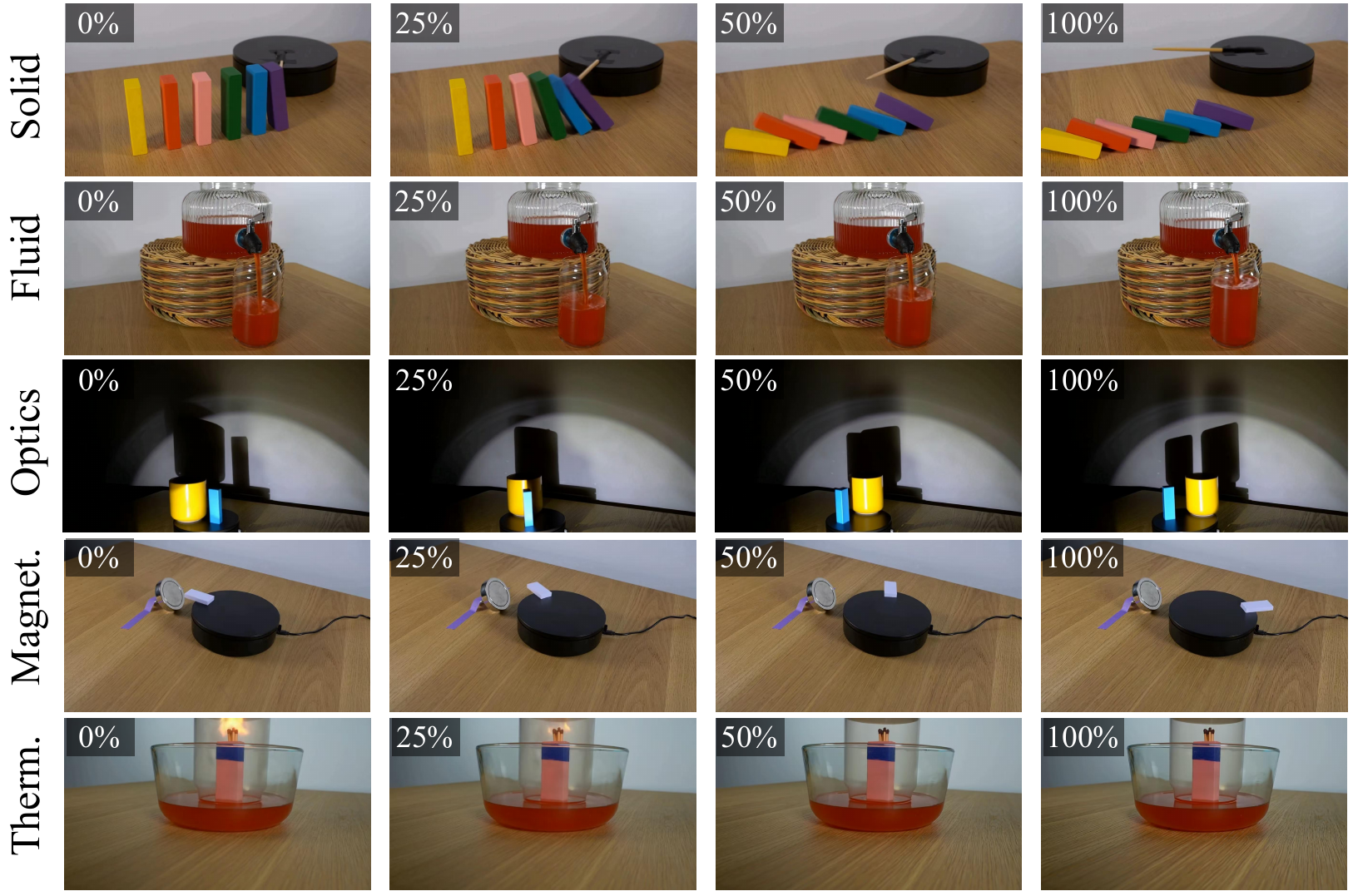}
    \vspace{-8pt}
    \caption{Visualization of generated videos.}
    \vspace{-8pt}
  \label{fig:vis}
  \vspace{-8pt}
\end{figure}

\noindent\textbf{Qualitative Analysis.}  
\cref{fig:vis} visualizes generated videos across five physical domains.  
For relatively simple physical dynamics, the VGM produces results that closely align with real-world physics, supporting the effectiveness of our method for physically consistent generation.

\begin{table}[t]
\centering
\caption{Performance across Iterative Loops on the Physics-IQ Benchmark. $\ast$ indicates partial refinement on incomplete prompts.}
\vspace{-8pt}
\setlength{\tabcolsep}{12pt}
\resizebox{0.44\textwidth}{!}{%
    \begin{tabular}{clc|c}
        \toprule
        \textbf{No.} & \textbf{Method} & \textbf{Infer. Steps} & \textbf{Physics-IQ Score ($\uparrow$) } \\
        \midrule
        1 & 1st Loop & 16 & 49.80 \\
        2 & 2nd Loop & 16 & 48.31$^{\ast}$ \\
        3 & 3rd Loop & 16 & 51.65 \\
        4 & 4th Loop & 16 & 52.92 \\
        \midrule
        5 & 1st Loop & 32 & 49.49 \\
        6 & 4th Loop & 32 & 49.15$^{\ast}$ \\
        \midrule
        7 & \multicolumn{2}{c|}{Ensemble \{1,2,5\}}  & 57.09 \\
        8 & \multicolumn{2}{c|}{Ensemble \{1,2,3,4,5,6\}} & \textbf{62.38} \\
        \bottomrule
    \end{tabular}
}
\label{tab:results}
\vspace{-10pt}
\end{table}

{
    \small
    \bibliographystyle{ieeenat_fullname}
    \bibliography{main}

\begin{thebibliography}{29}
\providecommand{\natexlab}[1]{#1}
\providecommand{\url}[1]{\texttt{#1}}
\expandafter\ifx\csname urlstyle\endcsname\relax
  \providecommand{\doi}[1]{doi: #1}\else
  \providecommand{\doi}{doi: \begingroup \urlstyle{rm}\Url}\fi

\bibitem[Achiam et~al.(2023)Achiam, Adler, Agarwal, Ahmad, Akkaya, Aleman, Almeida, Altenschmidt, Altman, Anadkat, et~al.]{GPT4}
Josh Achiam, Steven Adler, Sandhini Agarwal, Lama Ahmad, Ilge Akkaya, Florencia~Leoni Aleman, Diogo Almeida, Janko Altenschmidt, Sam Altman, Shyamal Anadkat, et~al.
\newblock Gpt-4 technical report.
\newblock \emph{arXiv preprint arXiv:2303.08774}, 2023.

\bibitem[Bai et~al.(2023)Bai, Bai, Chu, Cui, Dang, Deng, Fan, Ge, Han, Huang, et~al.]{Qwen}
Jinze Bai, Shuai Bai, Yunfei Chu, Zeyu Cui, Kai Dang, Xiaodong Deng, Yang Fan, Wenbin Ge, Yu Han, Fei Huang, et~al.
\newblock Qwen technical report.
\newblock \emph{arXiv preprint arXiv:2309.16609}, 2023.

\bibitem[Bansal et~al.(2024)Bansal, Lin, Xie, Zong, Yarom, Bitton, Jiang, Sun, Chang, and Grover]{challenge3}
Hritik Bansal, Zongyu Lin, Tianyi Xie, Zeshun Zong, Michal Yarom, Yonatan Bitton, Chenfanfu Jiang, Yizhou Sun, Kai-Wei Chang, and Aditya Grover.
\newblock Videophy: Evaluating physical commonsense for video generation.
\newblock \emph{arXiv preprint arXiv:2406.03520}, 2024.

\bibitem[Bansal et~al.(2025)Bansal, Peng, Bitton, Goldenberg, Grover, and Chang]{challenge4}
Hritik Bansal, Clark Peng, Yonatan Bitton, Roman Goldenberg, Aditya Grover, and Kai-Wei Chang.
\newblock Videophy-2: A challenging action-centric physical commonsense evaluation in video generation.
\newblock \emph{arXiv preprint arXiv:2503.06800}, 2025.

\bibitem[Bar-Tal et~al.(2024)Bar-Tal, Chefer, Tov, Herrmann, Paiss, Zada, Ephrat, Hur, Liu, Raj, et~al.]{Lumiere}
Omer Bar-Tal, Hila Chefer, Omer Tov, Charles Herrmann, Roni Paiss, Shiran Zada, Ariel Ephrat, Junhwa Hur, Guanghui Liu, Amit Raj, et~al.
\newblock Lumiere: A space-time diffusion model for video generation.
\newblock In \emph{SIGGRAPH Asia 2024}, pages 1--11, 2024.

\bibitem[Comanici et~al.(2025)Comanici, Bieber, Schaekermann, Pasupat, Sachdeva, Dhillon, Blistein, Ram, Zhang, Rosen, et~al.]{Gemini}
Gheorghe Comanici, Eric Bieber, Mike Schaekermann, Ice Pasupat, Noveen Sachdeva, Inderjit Dhillon, Marcel Blistein, Ori Ram, Dan Zhang, Evan Rosen, et~al.
\newblock Gemini 2.5: Pushing the frontier with advanced reasoning, multimodality, long context, and next generation agentic capabilities.
\newblock \emph{arXiv preprint arXiv:2507.06261}, 2025.

\bibitem[Dai et~al.(2025)Dai, Xu, Wen, Liu, and Huang]{dai2025exploring}
Siran Dai, Qianqian Xu, Peisong Wen, Yang Liu, and Qingming Huang.
\newblock Exploring structural degradation in dense representations for self-supervised learning.
\newblock \emph{arXiv preprint arXiv:2510.17299}, 2025.

\bibitem[Floridi and Chiriatti(2020)]{GPT3}
Luciano Floridi and Massimo Chiriatti.
\newblock Gpt-3: Its nature, scope, limits, and consequences.
\newblock \emph{Minds and machines}, 30\penalty0 (4):\penalty0 681--694, 2020.

\bibitem[Hong et~al.(2022)Hong, Ding, Zheng, Liu, and Tang]{pretrain1}
Wenyi Hong, Ming Ding, Wendi Zheng, Xinghan Liu, and Jie Tang.
\newblock Cogvideo: Large-scale pretraining for text-to-video generation via transformers.
\newblock \emph{arXiv preprint arXiv:2205.15868}, 2022.

\bibitem[Hwang et~al.(2025)Hwang, Jang, Kim, Park, and Choo]{distill2}
Sungwon Hwang, Hyojin Jang, Kinam Kim, Minho Park, and Jaegul Choo.
\newblock Cross-frame representation alignment for fine-tuning video diffusion models.
\newblock \emph{arXiv preprint arXiv:2506.09229}, 2025.

\bibitem[Kang et~al.(2024)Kang, Yue, Lu, Lin, Zhao, Wang, Huang, and Feng]{challenge2}
Bingyi Kang, Yang Yue, Rui Lu, Zhijie Lin, Yang Zhao, Kaixin Wang, Gao Huang, and Jiashi Feng.
\newblock How far is video generation from world model: A physical law perspective.
\newblock \emph{arXiv preprint arXiv:2411.02385}, 2024.

\bibitem[Kondratyuk et~al.(2024)Kondratyuk, Yu, Gu, Lezama, Huang, Schindler, Hornung, Birodkar, Yan, Chiu, et~al.]{VideoPoet}
Dan Kondratyuk, Lijun Yu, Xiuye Gu, Jose Lezama, Jonathan Huang, Grant Schindler, Rachel Hornung, Vighnesh Birodkar, Jimmy Yan, Ming-Chang Chiu, et~al.
\newblock Videopoet: A large language model for zero-shot video generation.
\newblock In \emph{International Conference on Machine Learning}, pages 25105--25124. PMLR, 2024.

\bibitem[Lin et~al.(2025)Lin, Wang, Wang, Wang, Dai, Ding, Wang, Zuo, Sang, Huang, et~al.]{challenge1}
Minghui Lin, Xiang Wang, Yishan Wang, Shu Wang, Fengqi Dai, Pengxiang Ding, Cunxiang Wang, Zhengrong Zuo, Nong Sang, Siteng Huang, et~al.
\newblock Exploring the evolution of physics cognition in video generation: A survey.
\newblock \emph{arXiv preprint arXiv:2503.21765}, 2025.

\bibitem[Liu et~al.(2024{\natexlab{a}})Liu, Feng, Xue, Wang, Wu, Lu, Zhao, Deng, Zhang, Ruan, et~al.]{Deepseek}
Aixin Liu, Bei Feng, Bing Xue, Bingxuan Wang, Bochao Wu, Chengda Lu, Chenggang Zhao, Chengqi Deng, Chenyu Zhang, Chong Ruan, et~al.
\newblock Deepseek-v3 technical report.
\newblock \emph{arXiv preprint arXiv:2412.19437}, 2024{\natexlab{a}}.

\bibitem[Liu et~al.(2024{\natexlab{b}})Liu, Xu, Wen, Dai, and Huang]{HAP_VR}
Yang Liu, Qianqian Xu, Peisong Wen, Siran Dai, and Qingming Huang.
\newblock Not all pairs are equal: Hierarchical learning for average-precision-oriented video retrieval.
\newblock In \emph{ACM International Conference on Multimedia}, pages 3828--3837, 2024{\natexlab{b}}.

\bibitem[Liu et~al.(2024{\natexlab{c}})Liu, Zhang, Li, Yan, Gao, Chen, Yuan, Huang, Sun, Gao, et~al.]{Sora}
Yixin Liu, Kai Zhang, Yuan Li, Zhiling Yan, Chujie Gao, Ruoxi Chen, Zhengqing Yuan, Yue Huang, Hanchi Sun, Jianfeng Gao, et~al.
\newblock Sora: A review on background, technology, limitations, and opportunities of large vision models.
\newblock \emph{arXiv preprint arXiv:2402.17177}, 2024{\natexlab{c}}.

\bibitem[Liu et~al.(2025)Liu, Xu, Wen, Dai, and Huang]{tcore}
Yang Liu, Qianqian Xu, Peisong Wen, Siran Dai, and Qingming Huang.
\newblock When the future becomes the past: Taming temporal correspondence for self-supervised video representation learning.
\newblock In \emph{IEEE/CVF Conference on Computer Vision and Pattern Recognition}, pages 24033--24044, 2025.

\bibitem[Motamed et~al.(2025{\natexlab{a}})Motamed, Chen, Gool, and Laina]{challenge5}
Saman Motamed, Minghao Chen, Luc~Van Gool, and Iro Laina.
\newblock Travl: A recipe for making video-language models better judges of physics implausibility, 2025{\natexlab{a}}.

\bibitem[Motamed et~al.(2025{\natexlab{b}})Motamed, Culp, Swersky, Jaini, and Geirhos]{PhyIQ}
Saman Motamed, Laura Culp, Kevin Swersky, Priyank Jaini, and Robert Geirhos.
\newblock Do generative video models understand physical principles?
\newblock \emph{arXiv preprint arXiv:2501.09038}, 2025{\natexlab{b}}.

\bibitem[Oquab et~al.(2023)Oquab, Darcet, Moutakanni, Vo, Szafraniec, Khalidov, Fernandez, Haziza, Massa, El-Nouby, et~al.]{dinov2}
Maxime Oquab, Timoth{\'e}e Darcet, Th{\'e}o Moutakanni, Huy Vo, Marc Szafraniec, Vasil Khalidov, Pierre Fernandez, Daniel Haziza, Francisco Massa, Alaaeldin El-Nouby, et~al.
\newblock Dinov2: Learning robust visual features without supervision.
\newblock \emph{arXiv preprint arXiv:2304.07193}, 2023.

\bibitem[Paszke et~al.(2019)Paszke, Gross, Massa, Lerer, Bradbury, Chanan, Killeen, Lin, Gimelshein, Antiga, et~al.]{Pytorch}
Adam Paszke, Sam Gross, Francisco Massa, Adam Lerer, James Bradbury, Gregory Chanan, Trevor Killeen, Zeming Lin, Natalia Gimelshein, Luca Antiga, et~al.
\newblock Pytorch: An imperative style, high-performance deep learning library.
\newblock \emph{Advances in neural information processing systems}, 32, 2019.

\bibitem[Sim{\'e}oni et~al.(2025)Sim{\'e}oni, Vo, Seitzer, Baldassarre, Oquab, Jose, Khalidov, Szafraniec, Yi, Ramamonjisoa, et~al.]{dinov3}
Oriane Sim{\'e}oni, Huy~V Vo, Maximilian Seitzer, Federico Baldassarre, Maxime Oquab, Cijo Jose, Vasil Khalidov, Marc Szafraniec, Seungeun Yi, Micha{\"e}l Ramamonjisoa, et~al.
\newblock Dinov3.
\newblock \emph{arXiv preprint arXiv:2508.10104}, 2025.

\bibitem[Teng et~al.(2025)Teng, Jia, Sun, Li, Li, Tang, Han, Zhang, Zhang, Luo, et~al.]{MAGI1}
Hansi Teng, Hongyu Jia, Lei Sun, Lingzhi Li, Maolin Li, Mingqiu Tang, Shuai Han, Tianning Zhang, WQ Zhang, Weifeng Luo, et~al.
\newblock Magi-1: Autoregressive video generation at scale.
\newblock \emph{arXiv preprint arXiv:2505.13211}, 2025.

\bibitem[Wang et~al.(2025)Wang, Ma, Cao, Zheng, Zhang, Feng, Liu, Ma, Cheng, Leng, et~al.]{prompt3}
Jing Wang, Ao Ma, Ke Cao, Jun Zheng, Zhanjie Zhang, Jiasong Feng, Shanyuan Liu, Yuhang Ma, Bo Cheng, Dawei Leng, et~al.
\newblock Wisa: World simulator assistant for physics-aware text-to-video generation.
\newblock \emph{arXiv preprint arXiv:2503.08153}, 2025.

\bibitem[Wen et~al.(2025)Wen, Xu, Dai, Cong, and Huang]{wen2025semantic}
Peisong Wen, Qianqian Xu, Siran Dai, Runmin Cong, and Qingming Huang.
\newblock Semantic concentration for self-supervised dense representations learning.
\newblock \emph{IEEE Transactions on Pattern Analysis and Machine Intelligence}, 2025.

\bibitem[Xue et~al.(2025)Xue, Yin, Yang, and Gao]{prompt1}
Qiyao Xue, Xiangyu Yin, Boyuan Yang, and Wei Gao.
\newblock Phyt2v: Llm-guided iterative self-refinement for physics-grounded text-to-video generation.
\newblock In \emph{IEEE/CVF Conference on Computer Vision and Pattern Recognition}, pages 18826--18836, 2025.

\bibitem[Yang et~al.(2024)Yang, Teng, Zheng, Ding, Huang, Xu, Yang, Hong, Zhang, Feng, et~al.]{pretrain2}
Zhuoyi Yang, Jiayan Teng, Wendi Zheng, Ming Ding, Shiyu Huang, Jiazheng Xu, Yuanming Yang, Wenyi Hong, Xiaohan Zhang, Guanyu Feng, et~al.
\newblock Cogvideox: Text-to-video diffusion models with an expert transformer.
\newblock \emph{arXiv preprint arXiv:2408.06072}, 2024.

\bibitem[Zhang et~al.(2025{\natexlab{a}})Zhang, Xiao, Mei, Xu, and Patel]{prompt2}
Ke Zhang, Cihan Xiao, Yiqun Mei, Jiacong Xu, and Vishal~M Patel.
\newblock Think before you diffuse: Llms-guided physics-aware video generation.
\newblock \emph{arXiv preprint arXiv:2505.21653}, 2025{\natexlab{a}}.

\bibitem[Zhang et~al.(2025{\natexlab{b}})Zhang, Liao, Zhang, Meng, Wan, Yan, and Cheng]{distill1}
Xiangdong Zhang, Jiaqi Liao, Shaofeng Zhang, Fanqing Meng, Xiangpeng Wan, Junchi Yan, and Yu Cheng.
\newblock Videorepa: Learning physics for video generation through relational alignment with foundation models.
\newblock \emph{arXiv preprint arXiv:2505.23656}, 2025{\natexlab{b}}.

\end{thebibliography}
}

\end{document}